\documentclass[twoside,leqno,twocolumn]{article}

\usepackage[letterpaper]{geometry}

\usepackage{booktabs}

\usepackage{hyperref}

\usepackage{ltexpprt}

\usepackage{algorithm}
\usepackage{algorithmic}

\usepackage{multirow}
\usepackage{bm}
\usepackage{enumitem}

\usepackage{amsmath}
\usepackage{amssymb}
\usepackage{mathtools}

\newcommand{\mat}[1]{\mathbf{#1}} 
\newcommand{\ve}[1]{\mathbf{#1}} 
\newcommand{\ten}[1]{\bm{\mathcal{#1}}}
\newcommand{\xhdr}[1]{{\noindent\bfseries #1}.}

\usepackage{caption}
\usepackage{subcaption}

\usepackage{tikz}

\usepackage{pgfplots}
\usepackage{xcolor}
\definecolor{bigbird_color}{RGB}{46,139,87}
\definecolor{longformer_color}{RGB}{107,143,179}
\definecolor{hipool_color}{RGB}{161,116,161}
\definecolor{hipool_bar}{RGB}{147,112,219}
\definecolor{longformer_bar}{RGB}{143,188,143}
\definecolor{bar_color}{RGB}{70,130,180}

\begin{document}

\newcommand\relatedversion{}

\title{\Large Efficient High-Resolution Time Series Classification via \\Attention Kronecker Decomposition\relatedversion}
\author{
Aosong Feng \thanks{Yale University}\and
Jialin Chen \footnotemark[1] \and
Juan Garza \thanks{The University of Texas Rio Grande Valley} \and
Brooklyn Berry \footnotemark[2] \and
Francisco Salazar \footnotemark[2]  \and
Yifeng Gao \footnotemark[2]  \and
Rex Ying \footnotemark[1] \and
Leandros Tassiulas \footnotemark[1]
}

\date{}

\maketitle


\fancyfoot[R]{\scriptsize{Copyright \textcopyright\ 20XX by SIAM\\
Unauthorized reproduction of this article is prohibited}}





\begin{abstract} 
The high-resolution time series classification problem is essential due to the increasing availability of detailed temporal data in various domains.
To tackle this challenge effectively, it is imperative that the state-of-the-art attention model is scalable to accommodate the growing sequence lengths typically encountered in high-resolution time series data, while also demonstrating robustness in handling the inherent noise prevalent in such datasets.
To address this, we propose to hierarchically encode the long time series into multiple levels based on the interaction ranges.
By capturing relationships at different levels, we can build more robust, expressive, and efficient models that are capable of capturing both short-term fluctuations and long-term trends in the data.
We then propose a new time series transformer backbone (\textbf{KronTime}) by introducing Kronecker-decomposed attention to process such multi-level time series, which sequentially calculates attention from the lower level to the upper level.
Experiments on four long time series datasets demonstrate superior classification results with improved efficiency compared to baseline methods.

\end{abstract}

\section{Introduction}

Multivariate Time Series Classification (MTSC) problem is one of the most essential time series data mining tasks that have a great impact in various fields such as manufacturing\cite{fan2020defective,mellah2022early}, astronomy\cite{wilson1994binary,matijevivc2012kepler, bassi2021classification}, and entomology\cite{yin2021lightweight,akter2020mosquito,alar2021accurate}. With the advancement of sensor technique, high-resolution and long-term passive monitoring time series\cite{metcalf2022optimizing,thuy2021segmentation} has become increasingly available. 

However, unlike traditional MTSC problems, the considerably long time series gathered can pose significant challenges in the classification task. The increasing time series length can impact the difficulty of the MTSC problem in two folds. First, the large length significantly increases the computation cost in a wide range of classification models such as Transformer based model, similarity comparison-based model\cite{jeong2011weighted}, shapelet-based model\cite{karlsson2016generalized}, and bag-of-patterns based approach\cite{senin2013sax}. Furthermore, such issues are further amplified in the era of parameter-intensive deep learning models. For example, a Transformer\cite{wolf2020transformers} which needs memory and computation cost quadratic to the length will be difficult to use in long sequence MTSC tasks. Second, unlike image and video, the time series can potentially become highly noisy in high-resolution and passive monitoring applications, as the sensor will be sensitive and a large amount of noise and distortion may be included. Such a low Signal-to-Noise ratio across the entire time series requires the classification model to have the capability to accurately capture useful information across a long range of time and map the data into low dimensions without intervening by the presence of noises. Therefore, an effective deep learning model that can model and accurately model data behavior and dependency across long-range is necessary for addressing long-sequence MTSC tasks.

In this paper, to overcome the these challenges, we propose to hierarchically encode time series by considering multi-level interactions.
As shown in Figure \ref{fig::model}(a), each level of time series encodings defines time step correlations with a certain distance, gradually covering short-range to long-range interactions.
Such hierarchical encoding alleviates the problem of short-range noisy patterns by including upper-level global information for high-resolution time series processing.
Besides, the effective time series length is much less than the total length of the original time series and therefore avoids the quadratic computation costs.
To accommodate the hierarchical encoding of the time series in the transformer model, we propose to decompose the original attention matrix using Kroncker decomposition according to the defined hierarchy and name the resulting model \textbf{KronTime}.
Given finite attention window size and memory budget, KronTime captures correlations along different dimensions and therefore includes hierarchical multi-hop token interactions in the original sequence at multiple scales.
Experiments on long-sequence time series classification show superior classification results of KronTime compared to state-of-the-art attention and convolution-based models, with improved running time and memory usage.

\section{Related Work}
\xhdr{Time Series Classification}
Traditional machine learning techniques like dynamic time warping (DTW) \cite{jeong2011weighted}, hierarchical vote collective of transformation-based ensembles (HIVE-COTE) \cite{lines2016hive}, and proximity forest \cite{lucas2019proximity} have long been prevalent in time series classification. However, they often fall short in terms of both efficiency and accuracy, particularly as datasets grow in size and complexity. To address the challenges of multivariate time series classification (MTSC), the field has witnessed a surge in the development of sophisticated deep learning models. Convolution-based architectures such as ResNet \cite{wang2017time}, InceptionTime \cite{ismail2020inceptiontime}, and TimesNet \cite{wu2023timesnet} have showcased their ability to capture local temporal patterns with high accuracy. Meanwhile, recurrent-based models \cite{rnn1, rnn2} excel in modeling sequential information but struggle with longer sequences. Transformer-based approaches~\cite{patchtst, liu2021gated, yuan2020self} recently have emerged as powerful tools for capturing complex long-range dependencies and cross-variate interactions, due to their effective utilization of the self-attention mechanism \cite{vaswani2017attention}. Additionally, multi-resolution strategies \cite{cui2016multi, chen2021multi, qian2020dynamic} continue to be instrumental in enhancing the performance of these advanced models. Nonetheless, challenges persist, including issues of model efficiency and the handling of hierarchical data structures, especially when handling long time series, highlighting the need for further research and innovation in the field.

\xhdr{Efficient Transformer for Time Series} 
Transformers, originally prominent in natural language processing (NLP) \cite{wolf2020transformers, kalyan2021ammus, tunstall2022natural}, computer vision (CV) \cite{liu2021swin, han2022survey}, and speech recognition \cite{dong2018speech, gulati2020conformer}, have recently attracted attention in addressing multivariate time series challenges. Several Transformer-based models have emerged, promising improved performance and enhanced model efficiency. For instance, LogTrans \cite{li2019enhancing} introduces LogSparse attention, while Informer \cite{zhou2021informer} proposes ProbSparse self-attention, both achieving logarithmic complexity concerning the time series length. Pyraformer \cite{liu2021pyraformer} incorporates a pyramidal attention module to efficiently capture temporal dependencies, while FEDformer \cite{zhou2022fedformer} leverages sparse representation in the frequency domain, presenting a frequency-enhanced Transformer with linear complexity. However, these advancements in self-attention efficiency may come at the cost of reduced expressiveness, leading to sub-optimal modeling performance. Additionally, these models often lack the ability to discern hierarchical data structures and capture multi-resolution information, both critical aspects in comprehensive time series analysis. Further research is warranted to address these limitations effectively.

\section{Methodology}
\begin{figure*}[tp]
  \vspace{-1pt}
  \centering \includegraphics[width=1.0\linewidth]{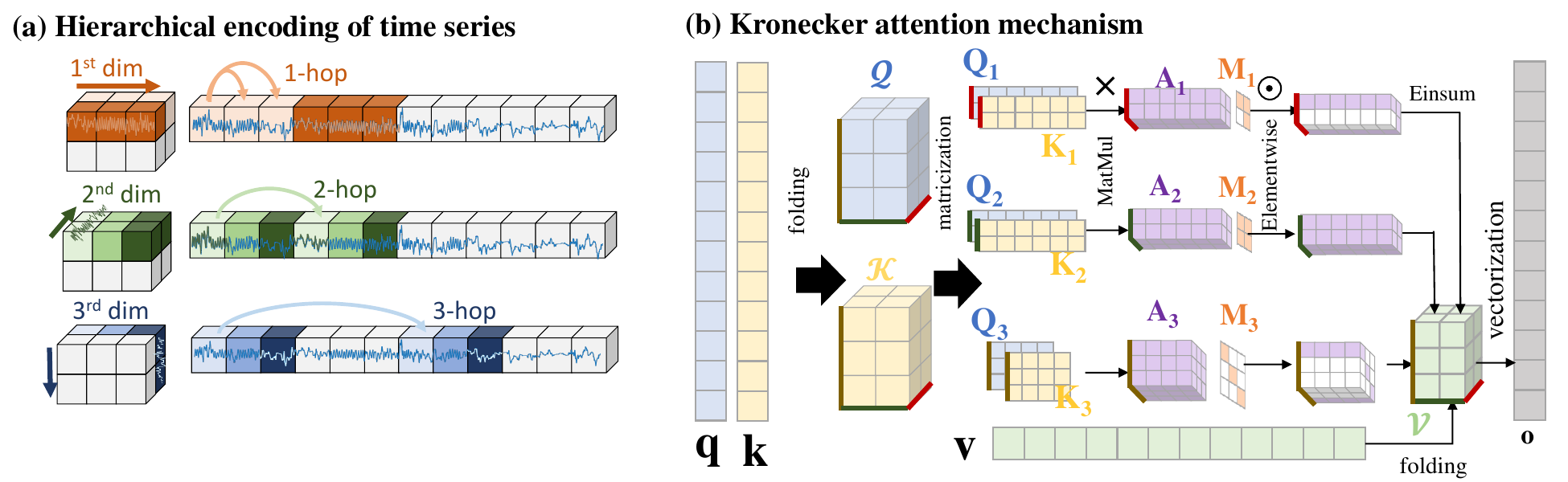}
  \caption{
  (a) The long time series after patchified can be decomposed into multiple levels. The first, second, and third level encodes adjacent, mid-range, and long-range global information, respectively
  (b) Input sequences $\ve{q}, \ve{k}, \ve{v}$ are first tensorized into $\ten{Q}, \ten{K}, \ten{V}$.
  Each row in the middle represents the attention along one matching dimension of tensors, and all dimensions except the matching dimension of $\ten{Q}$ and $\ten{K}$ are flattened.
  The result from each row is used to sequentially update the value tensor $\ten{V}$.
  }
  \label{fig::model}
  \vspace{-5pt}
\end{figure*}

Long Time series often exhibit complex structures and dependencies that span multiple scales and granularities.
By encoding the data hierarchically, we can represent these relationships more structured and interpretably. 
Kronecker decomposition, in particular, allows us to decompose the time series into a series of hierarchical tensors, where each level captures increasingly abstract representations of the underlying patterns.
This hierarchical approach enables the model to learn representations at different levels of temporal granularity, from fine-grained details to higher-level trends and patterns.

\subsection{Hierarchical Time Series Encoding}
Given input time series with $n$ time steps, the transformer attention layer take inputs $\ve{x}\in\mathbb{R}^{n\times d}$, query $\ve{q}$, key $\ve{k}$ and value $\ve{v}$ which are derived from linear projections as $\ve{q} = \mat{xW_q}, \ve{k} = \mat{xW_k}, \ve{v}=\mat{xW_v}$.
The attention process can then be written as 
\begin{equation}
\label{eq::softmax}
\begin{split}
\mat{A}=\text{softmax}\left(\dfrac{\mat{qk}^\intercal}{\sqrt{d}} \right), \ve{o} = \mat{A}\ve{v}
\end{split}
\end{equation}
where $\text{softmax}$ denotes the row-wise softmax normalization, $\ve{o}$ is the updated value vector as output.

To capture the time series at different levels and avoid the quadratic computation costs in the long sequence scenario, we propose to reshape the original sequence into a compact tensor as shown in Figure \ref{fig::model}(a), with each dimension encoding the information of a certain level.
Specifically, we reshape $\ve{q}, \ve{k}, \ve{v}$ into order-$m$ tensors $\ten{Q},\ten{K},\ten{V} \in \mathbb{R}^{n_1\times ...n_m}$ $\prod_{i=1}^{m}n_i=n$, representing $m$-level decomposition.
We then consider the attention interactions only at the same level,
in other words, to update the $i$-th dimension of value tensor $\ten{V}$, we calculate the attention between the $i$-th dimension of $\ten{Q}$ and $\ten{K}$ while treating other dimensions as batch dimensions.
The update of $\ten{V}$ can then be achieved by sequential updating from the first to the last level (irrelevant to the order of updating).

\subsection{Attention Decomposition}
To model the tensor interaction between $\ten{Q},\ten{K}, \ten{V}$, we propose attention Kronecker decomposition which sequentially models the interactions from the short range to the long range. 
Since the input length along each dimension is much smaller than the entire sequence, such decomposed attention can work with much less context window budget.
The attention process in Equation \ref{eq::softmax} can be decomposed with Kronecker decomposition as 
\begin{equation}
\label{eq::kron_decom}
\begin{split}
\mat{A} = \otimes_{i=1}^{m}\mat{A}_i, \ \   \mat{O} = \mat{A} \mat{V}, 
\end{split}
\end{equation}
where $\mat{A}_i$ models the $i$-th level token interaction.
Considering the property of mixed Kronecker-matrix product, the above value tensor updating can be sequential, irrespective of the updating order.
For efficient implementation, we adopt such sequential calculation of Equation \ref{eq::kron_decom} and model the $i$-th value updating as matrix multiplication
\begin{equation}
\label{eq::kron_decom}
\begin{split}
\mat{O}_i = \mat{A}_i \mat{V}_i
\end{split}
\end{equation}
where $\mat{V}_i\in\mathbb{R}^{n_i\times (n_1..n_{i-1}n_{i+1}..n_m)}$ is the mode-$i$ flattening of tensor $\ten{V}$, and $\mat{O}_i$ will be used as the value matrix in the next update.
Mode-$i$ flattening reshapes a tensor $\ten{T}\in\mathbb{R}^{n_1\times ..\times n_m}$ into matrix $\mat{T}_i\in\mathbb{R}^{(n_1..n_{i-1}n_{i+1}..n_m) \times n_i}$ which can be interpreted as batching $n_1..n_{i-1}n_{i+1}..n_m$ vectors.

The attention matrix $\mat{A}_i$ is used to model the $i$-th level correlation between query and key
\begin{equation}
\label{eq::qk}
\begin{split}
\mat{A}_i=\text{softmax}\left(\dfrac{\mat{Q}_i\mat{K}_i^\intercal}{ \sqrt{d}} \right),
\end{split}
\end{equation}
where $\mat{Q}_i, \mat{K}_i$ is the mode-$i$ flattening of the hierarchical time series encoding tensors $\ten{Q}, \ten{K}$.

The sequential attention update is visualized in Figure \ref{fig::model} (b).
For the $i$-th dimension update,  $\ten{Q},\ten{K}$ are matricized by mode-$i$ flattening.
The resulting batched attention matrix is then used to update the $i$-th dimension of the value tensor.
The computational complexity is decreased from $\mathcal{O}(n^2)$ with full attention to $\mathcal{O}(nlogn)$.
We summarize the forward pass of Kronecker-decomposed attention in Algorithm \ref{alg::forward}.
\begin{algorithm}[h]
   \caption{KronTime forward}
   \label{alg::forward}
\begin{algorithmic}
   \STATE {\bfseries Input:} $\ten{Q},\ten{K}, \ten{V} \in \mathbb{R}^{n_1\times .. \times n_m}$
 
   \STATE Initialize $\ten{O} = \ten{V}$.
   \FOR{$i=0$ {\bfseries to} $m-1$}
   \STATE Mode-i flattening $\ten{Q},\ten{K}$ into $\mat{Q}_i,\mat{K}_i$
   \STATE $\mat{A}_i=\text{softmax}(\mat{Q}_i\mat{K}_i^\intercal/{ \sqrt{d}} \circ \mat{M}_i)$
   \STATE Mode-i flattening $\ten{O}$ into $\mat{O}_i$
   \STATE Value updates: $\mat{O}_i = \mat{A}_i \mat{O}_i$
   \STATE Mode-i folding $ \mat{O}_i$ into $\ten{O}$
   \ENDFOR
   \STATE Vectorize $\ten{O}$ to $\ve{o}$
   \STATE Return $\ve{o}$
\end{algorithmic}
\end{algorithm}

\subsection{Model Implementations} We choose SOTA time series transformers PatchTST \cite{nie2022time} as the backbone, and replace the full attention in the attention layer with the Kronecker decomposed attention as discussed above.
For classification applications, we add an additional classification head with linear projections which takes the last hidden states as inputs.

\section{Experiment Evaluation}

\begin{table*}[ht]
\caption{Time Series Classification Performance Comparison}
\label{tab::acc}
\centering
\begin{tabular}{l cccc} \toprule
 \cmidrule{2-5}
\multicolumn{1}{l}{Model} & \multicolumn{1}{l}{BinaryHeartbeat} & \multicolumn{1}{l}{EigenWorms} & \multicolumn{1}{l}{FaultDetectionA} & \multicolumn{1}{l}{CatsDogs} \\  \cmidrule{1-5}
1-NN (ED)      & 0.585                                & 0.500                           & 0.460                           & 0.420                                 \\
ResNet18       & 0.630                                & 0.420                           & 0.990                           & 0.570                                 \\
RandomShapelet & 0.585                                & 0.692                           & N/A                           & 0.606                        \\
DLinear        & 0.658                                & 0.423                           & 0.532                           & 0.516                                 \\
TCN            & \textbf{0.707}                                & \textbf{0.769}                  & 0.986                           & 0.575                                 \\
PatchTST       & \textbf{0.707}                       & 0.692                           &           0.991        & 0.810                        \\  
KronTime & \textbf{0.707}                      & \textbf{0.769}                           & \textbf{0.996}             & \textbf{0.844}                                 \\    \bottomrule
\end{tabular}
\end{table*}

\subsection{Baselines}

The proposed method will be compared with the following four baselines.

\textbf{1-NN (ED)}\cite{ruiz2021great}: 1-NN (Euclidean Distance) is a simple yet effective algorithm for classification tasks, comparing each test instance to all training instances based on Euclidean distance.

\textbf{ResNet18}\cite{ismail2019deep}: ResNet18 is a widely used convolutional neural network architecture known for its depth and skip connections, which facilitate the effective learning of complex features in image data.

\textbf{RandomShapelet}\cite{karlsson2016generalized}: RandomShapelet is a model leveraging shapelet transform for time series classification, extracting discriminative sub-series (shapelets) from the data to distinguish between different classes.

\textbf{DLinear}~\cite{zeng2023transformers}: DLinear proposes to utilize a single linear layer that aggregates information from all history records. DLinear avoids temporal information loss caused by the nature of the permutation-invariant self-attention mechanism in transformer-based architectures and demonstrates promising performance on time series analysis.

\textbf{Temporal Convolutional Network}~\cite{bai2018empirical}: Temporal Convolutional Network (TCN) is a type of neural network specifically designed for processing temporal sequences, utilizing convolutional layers with dilated kernels to capture long-range dependencies and temporal patterns efficiently.

\textbf{PatchTST}: PatchTST proposed segmentation of time series into subseries-level patches as input tokens to Transformer.

\subsection{Benchmark Dataset}

From the UEA Time Series Classification Archive, we use the following fine-resolution time series datasets that each have lengths greater than 10,000 and have a sufficient number of samples. As a result, we use the following datasets to evaluate the proposed method: 

\textbf{BinaryHeartbeat}: BinaryHeartbeat dataset comprises heart sound recordings collected from healthy individuals and patients with cardiac conditions. The dataset contains 409 instances, recorded at 2,000Hz, with a time series length of 18,530. The classes include 110 normal and 299 abnormal instances. \\ 

\textbf{EigenWorms}: EigenWorms dataset focuses on Caenorhabditis elegans, a roundworm used extensively in genetics research as a model organism. The dataset collected the worm's movements represented by combinations of six base shapes, known as eigenworms. The dataset consists of 257 samples. Each samples consists of 17,891 in length. The goal is to classify individual worms as either wild-type (N2 reference strain) or one of four mutant types: goa-1, unc-1, unc-38, and unc-63. \\ 

\textbf{FaultDetectionA}: FaultDetection dataset contains 13,640 time series. Each one responding to the signal recorded from rolling bearing monitor system. There are total three types of classes. Undamaged rolling bearing system, inner damaged rolling bearing system, and outer damaged rolling bearing system. Each record consists of 5,120 in length.

\textbf{CatsDogs}: CatsDogs dataset consists of total 277 samples sound records from cats and dogs. Each recording consists of 14,773 in length\\

\subsection{Experiment Setting}

For each dataset, we uses $80\%$ of data as training data, $10\%$ of data as validation data, and the rest $10\%$ of data as testing data. The performance is evaluated on the accuracy in the testing data as well as the efficiency of KronTime.
We run experiments with the same train/validation/test split on the four datasets. 
The final test accuracy is obtained using the checkpoint of the lowest validation loss, with the early-stop patience epochs 20. 
All experiments are performed on A6000 GPUs.

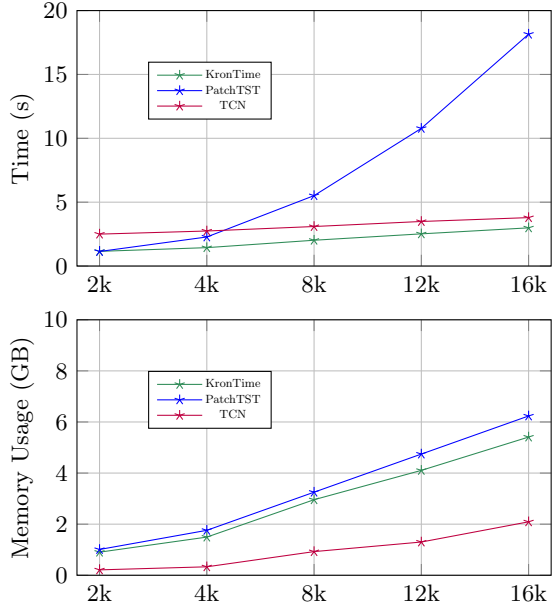
\begin{figure}[ptb]
\centering
\begin{subfigure}[b]{.98\linewidth}
\begin{tikzpicture}
\centering
\tikzstyle{every node}=[font=\small]
\begin{axis}[
    width  = 0.95\linewidth,
    height = 0.6\linewidth,
    ymin = 0, ymax = 20,
    symbolic x coords = {2k, 4k, 8k, 12k, 16k},
    ytick = {},
    ylabel = {Time (s)},
    ymajorgrids = true,
    xmajorgrids = true,
    y label style={at={(0.08,0.5)}},
    enlarge x limits={abs=0.3cm},
     legend style={
        at={(0.15,0.8)},
        anchor=north west,
        nodes={scale=0.5, transform shape},
        }
]
\addplot[color = bigbird_color, mark = star ]
    coordinates {
    (2k, 1.15)
    (4k, 1.44)
    (8k, 2.02)
    (12k, 2.52)
    (16k, 2.99)
    };
\addlegendentry{KronTime}

\addplot[color = blue, mark = star]
    coordinates {
    (2k, 1.14)
    (4k, 2.275)
    (8k, 5.513)
    (12k, 10.778)
    (16k, 18.151)
    };
\addlegendentry{PatchTST}

\addplot[color = purple, mark = star]
    coordinates {
    (2k, 2.4998)
    (4k, 2.7426)
    (8k, 3.0947)
    (12k, 3.4885)
    (16k, 3.7916)
    };
\addlegendentry{TCN}
\end{axis}
\end{tikzpicture}
\label{fig:tokenlength_amazon}

\end{subfigure}
\begin{subfigure}[b]{.98\linewidth}
\begin{tikzpicture}
\centering
\tikzstyle{every node}=[font=\small]
\begin{axis}[
    width  = 0.95\linewidth,
    height = 0.6\linewidth,
    ymin = 0, ymax = 10,
    symbolic x coords = {2k, 4k, 8k, 12k, 16k},
    ytick = {},
    ylabel = {Memory Usage (GB)},
    ymajorgrids = true,
    xmajorgrids = true,
    y label style={at={(0.08,0.5)}},
    enlarge x limits={abs=0.3cm},
     legend style={
        at={(0.15,0.8)},
        anchor=north west,
        nodes={scale=0.5, transform shape},
        }
]
\addplot[color = bigbird_color, mark = star ]
    coordinates {
    (2k, 0.904)
    (4k, 1.4938)
    (8k, 2.9575)
    (12k, 4.1055)
    (16k, 5.41)
    };
\addlegendentry{KronTime}

\addplot[color = blue, mark = star]
    coordinates {
    (2k, 1.009)
    (4k, 1.759)
    (8k, 3.252)
    (12k, 4.744)
    (16k, 6.235)
    };
\addlegendentry{PatchTST}

\addplot[color = purple, mark = star]
    coordinates {
    (2k, 0.2113)
    (4k, 0.3310)
    (8k, 0.9259)
    (12k, 1.3011)
    (16k, 2.0917)
    };
\addlegendentry{TCN}
\end{axis}
\end{tikzpicture}
\label{fig:tokenlength_ildc}
\end{subfigure}
\caption{Comparison of running time and GPU memory usage with different input lengths.}
\label{fig:efficiency}
\end{figure}

\subsection{Classification Performance Evaluation}
The classification result is shown in Table \ref{tab::acc}. KronTime achieves the same or superior classification accuracy compared to SOTA models. 
Notably, KronTime achieves such performance with improved efficiency as shown in Figure \ref{fig:efficiency}.
To make fair comparisons, we apply FlashAttention-2 to PatchTST to replace its original slow PyTorch implementation for memory-efficient calculations.
Results show that KronTime achieves around $0.3\times$ running time compared to PatchTST at length 16k and is comparable to the conventional convolution-based TCN.
Such advantage of improved running time from using attention decomposition is larger as input length grows.
Besides, the memory usage of PatchTST and KronTime stays relatively constant as input length grows, because of FlashAttention-2 and hierarchical decomposition, respectively.

\subsection{Parameter Testing}
\begin{figure}
\centering
\begin{subfigure}[b]{.98\linewidth}
\begin{tikzpicture}
\centering
\tikzstyle{every node}=[font=\small]
\begin{axis}[
    width  = 0.95\linewidth,
    height = 0.6\linewidth,
    ymin = 0.9, ymax = 1,
    xtick = {0, 20, 40, 60, 80, 100},
    ytick = {},
    xlabel = {Epochs},
    ylabel = {Accuracy},
    ymajorgrids = true,
    xmajorgrids = true,
    y label style={at={(0.05,0.5)}},
    enlarge x limits={abs=0.3cm},
     legend style={
        at={(0.65,0.6)},
        anchor=north west,
        nodes={scale=0.5, transform shape},
        }
]

\addplot[color = bigbird_color, mark = circle ]
    coordinates {
    (0, 0.7544247508049011)
    (1, 0.9461652040481567)
    (2, 0.9351032376289368)
    (3, 0.9151917099952698)
    (4, 0.9616519212722778)
    (5, 0.9631268382072449)
    (6, 0.9616519212722778)
    (7, 0.8709439635276794)
    (8, 0.9439527988433838)
    (9, 0.983775794506073)
    (10, 0.974926233291626)
    (11, 0.976401150226593)
    (12, 0.976401150226593)
    (13, 0.9734513163566589)
    (14, 0.9815633893013)
    (15, 0.9808259606361389)
    (16, 0.98525071144104)
    (17, 0.98525071144104)
    (18, 0.9845132827758789)
    (19, 0.983775794506073)
    (20, 0.9815633893013)
    (21, 0.985988199710846)
    (22, 0.983775794506073)
    (23, 0.98893803358078)
    (24, 0.980088472366333)
    (25, 0.991150438785553)
    (26, 0.976401150226593)
    (27, 0.991150438785553)
    (28, 0.9896755218505859)
    (29, 0.9904129505157471)
    (30, 0.9904129505157471)
    (31, 0.9904129505157471)
    (32, 0.98893803358078)
    (33, 0.98893803358078)
    (34, 0.985988199710846)
    (35, 0.9896755218505859)
    (36, 0.991150438785553)
    (37, 0.987463116645813)
    (38, 0.991150438785553)
    (39, 0.9867256879806519)
    (40, 0.9904129505157471)
    (41, 0.9904129505157471)
    (42, 0.991150438785553)
    (43, 0.9904129505157471)
    (44, 0.991150438785553)
    (45, 0.9918879270553589)
    (46, 0.9896755218505859)
    (47, 0.9904129505157471)
    (48, 0.9904129505157471)
    (49, 0.991150438785553)
    (50, 0.991150438785553)
    (51, 0.9918879270553589)
    (52, 0.9918879270553589)
    (53, 0.9904129505157471)
    (54, 0.9918879270553589)
    (55, 0.9904129505157471)
    (56, 0.9918879270553589)
    (57, 0.991150438785553)
    (58, 0.9918879270553589)
    (59, 0.991150438785553)
    (60, 0.991150438785553)
    (61, 0.991150438785553)
    (62, 0.9918879270553589)
    };
\addlegendentry{1 level}

\addplot[color = blue, mark = circle]
    coordinates {
(0, 0.8230088353157043)
(1, 0.9233038425445557)
(2, 0.9756637215614319)
(3, 0.9653392434120178)
(4, 0.9417403936386108)
(5, 0.9594395160675049)
(6, 0.98893803358078)
(7, 0.985988199710846)
(8, 0.9882006049156189)
(9, 0.9904129505157471)
(10, 0.9292035102844238)
(11, 0.98893803358078)
(12, 0.98525071144104)
(13, 0.987463116645813)
(14, 0.9896755218505859)
(15, 0.9867256879806519)
(16, 0.991150438785553)
(17, 0.9933628439903259)
(18, 0.9830383658409119)
(19, 0.9830383658409119)
(20, 0.9955751895904541)
(21, 0.983775794506073)
(22, 0.99262535572052)
(23, 0.9896755218505859)
(24, 0.994837760925293)
(25, 0.994837760925293)
(26, 0.9955751895904541)
(27, 0.994837760925293)
(28, 0.9955751895904541)
(29, 0.9955751895904541)
(30, 0.9955751895904541)
(31, 0.9955751895904541)
(32, 0.9918879270553589)
(33, 0.9955751895904541)
(34, 0.994837760925293)
(35, 0.994837760925293)
(36, 0.991150438785553)
(37, 0.994837760925293)
(38, 0.994837760925293)
(39, 0.99262535572052)
(40, 0.99631267786026)
(41, 0.994837760925293)
(42, 0.994837760925293)
(43, 0.9941002726554871)
(44, 0.98893803358078)
(45, 0.994837760925293)
(46, 0.99631267786026)
(47, 0.994837760925293)
(48, 0.99631267786026)
(49, 0.99631267786026)
(50, 0.994837760925293)
(51, 0.9955751895904541)
(52, 0.9955751895904541)
(53, 0.99631267786026)
(54, 0.99631267786026)
(55, 0.99631267786026)
(56, 0.99631267786026)
(57, 0.99631267786026)
(58, 0.99631267786026)
(59, 0.99631267786026)
(60, 0.99631267786026)
    };
    
\addlegendentry{2 levels}

\addplot[color = purple, mark = circle]
    coordinates {
(0, 0.9174041152000427)
(1, 0.8215339183807373)
(2, 0.8672566413879395)
(3, 0.9277285933494568)
(4, 0.9181416034698486)
(5, 0.9505899548530579)
(6, 0.9609144330024719)
(7, 0.9550147652626038)
(8, 0.9682890772819519)
(9, 0.9668141603469849)
(10, 0.9299409985542297)
(11, 0.9705014824867249)
(12, 0.9609144330024719)
(13, 0.9734513163566589)
(14, 0.9734513163566589)
(15, 0.9564896821975708)
(16, 0.9741888046264648)
(17, 0.974926233291626)
(18, 0.967551589012146)
(19, 0.971238911151886)
(20, 0.9705014824867249)
(21, 0.969763994216919)
(22, 0.976401150226593)
(23, 0.9741888046264648)
(24, 0.978613555431366)
(25, 0.980088472366333)
(26, 0.9815633893013)
(27, 0.976401150226593)
(28, 0.974926233291626)
(29, 0.9771386384963989)
(30, 0.980088472366333)
(31, 0.9815633893013)
(32, 0.980088472366333)
(33, 0.982300877571106)
(34, 0.9808259606361389)
(35, 0.9815633893013)
(36, 0.9808259606361389)
(37, 0.982300877571106)
(38, 0.983775794506073)
(39, 0.9808259606361389)
(40, 0.9815633893013)
(41, 0.9830383658409119)
(42, 0.983775794506073)
(43, 0.982300877571106)
(44, 0.9830383658409119)
(45, 0.9845132827758789)
(46, 0.983775794506073)
(47, 0.9830383658409119)
(48, 0.983775794506073)
(49, 0.945132827758789)
(50, 0.9845132827758789)
(51, 0.983775794506073)
(52, 0.98525071144104)
(53, 0.983775794506073)
(54, 0.983775794506073)
(55, 0.983775794506073)
(56, 0.9845132827758789)
(57, 0.9845132827758789)
(58, 0.983775794506073)
(59, 0.983775794506073)
(60, 0.9845132827758789)
(61, 0.983775794506073)
(62, 0.9830383658409119)
(63, 0.983775794506073)
(64, 0.983775794506073)
(65, 0.9830383658409119)
(66, 0.983775794506073)
(67, 0.983775794506073)
(68, 0.983775794506073)
(69, 0.9830383658409119)
(70, 0.983775794506073)
(71, 0.9845132827758789)
(72, 0.983775794506073)
(73, 0.9830383658409119)
(74, 0.982300877571106)
(75, 0.982300877571106)
(76, 0.9830383658409119)
(77, 0.9830383658409119)
(78, 0.983775794506073)
(79, 0.9845132827758789)
(80, 0.983775794506073)
    };
\addlegendentry{3 levels}

\addplot[color = orange, mark = circle]
    coordinates {
(0, 0.9166666865348816)
(1, 0.9218289256095886)
(2, 0.9336283206939697)
(3, 0.9550147652626038)
(4, 0.9269911646842957)
(5, 0.8340708017349243)
(6, 0.9410029649734497)
(7, 0.9653392434120178)
(8, 0.982300877571106)
(9, 0.9321534037590027)
(10, 0.9284660816192627)
(11, 0.9557521939277649)
(12, 0.967551589012146)
(13, 0.9771386384963989)
(14, 0.9771386384963989)
(15, 0.9734513163566589)
(16, 0.983775794506073)
(17, 0.98525071144104)
(18, 0.9734513163566589)
(19, 0.971238911151886)
(20, 0.9830383658409119)
(21, 0.9572271108627319)
(22, 0.9808259606361389)
(23, 0.967551589012146)
(24, 0.980088472366333)
(25, 0.9867256879806519)
(26, 0.9808259606361389)
(27, 0.98525071144104)
(28, 0.982300877571106)
(29, 0.982300877571106)
(30, 0.982300877571106)
(31, 0.98525071144104)
(32, 0.9808259606361389)
(33, 0.9808259606361389)
(34, 0.9808259606361389)
(35, 0.9793510437011719)
(36, 0.9830383658409119)
(37, 0.9808259606361389)
(38, 0.980088472366333)
(39, 0.9778761267662048)
(40, 0.983775794506073)
(41, 0.9830383658409119)
(42, 0.9830383658409119)
(43, 0.9815633893013)
(44, 0.9756637215614319)
(45, 0.9734513163566589)
    };
\addlegendentry{4 levels}

\end{axis}
\end{tikzpicture}
\label{fig:tokenlength_amazon}

\end{subfigure}
\begin{subfigure}[b]{.98\linewidth}
\input{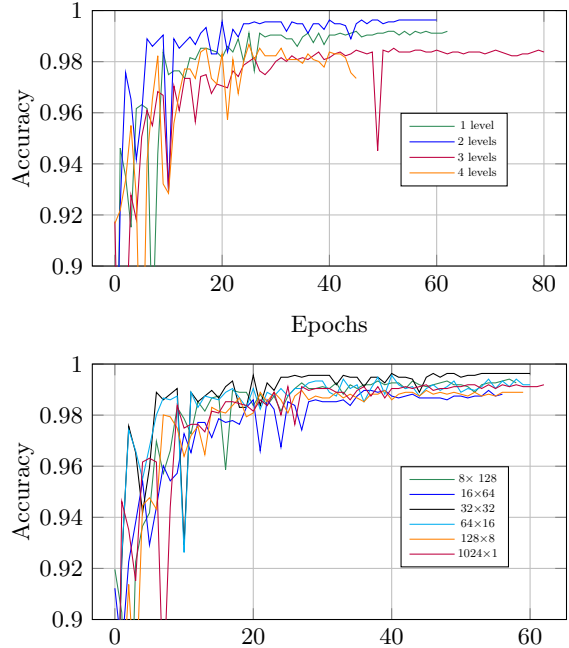}

\label{fig:tokenlength_ildc}
\end{subfigure}
\caption{The validation accuracy with different Kronecker decomposition strategies (upper: number of levels decomposed; lower: different decomposition with 2 levels) during the training phase.}
\label{fig:ablation}
\end{figure}

We next demonstrate the influence of the Kronecker decomposition by comparing the training curves (with training stop tolerance 20 epochs) under different decomposition strategies in KronTime. 
We perform such ablation studies on FaultDetectionA dataset with 1024 input length after tokenized.
We change the total number of levels decomposed and the size of each level while keeping other model and training hyperparameters unchanged for fair comparisons.
As shown in Figure \ref{fig:ablation}(a), the training with 2-level decomposition ($16\times 16$) converges to the higher validation accuracy with faster speed, compared to no decomposition (1-level decomposition), 3-level decomposition ($16\times 16\times 4$), and 4-level decomposition ($16\times 4\times4\times 4$).
This indicates that for FaultDetectionA dataset with 1024 length, 2-level Kronecker decomposition leads to the optimal result.
We then compare different decomposition strategies with 2 levels, and results in Figure \ref{fig:ablation}(b) show that decompositing the $1024$ sequence into $32\times 32$ achieves superior results compared to other decompositions for this dataset.

\section{Conclusion}

The high-resolution time series classification problem is essential due to the increasing availability of high-fidelity time series data. The growth of such high-resolution data will pose challenges in designing classification models. To tackle this challenge, we proposed Kronecker-decomposed attention (KronTime) to extract features from time series over 10,000 lengths effectively. The experiment demonstrates that KronTime can achieve superior classification results with improved efficiency compared to baselines.

\bibliography{transformer}
\bibliographystyle{plain}
\end{document}